\def\cvprPaperID{0000}
\def\confName{CVPR}
\def\confYear{2025}

\def\authorBlock{

    Haotian Dong$^{1}$ \enskip \enskip 
    Xin Wang$^{2}$ \enskip\enskip 
    Di Lin$^{1}$\footnotemark[2] \enskip\enskip 
    Yipeng Wu$^{1}$ \enskip\enskip
    Qin Chen$^{1}$ \enskip\enskip \\
    Ruonan Liu$^{3}$ \enskip\enskip
    Kairui Yang$^{1}$ \enskip\enskip
    Ping Li$^{2}$ \enskip\enskip
    Qing Guo$^{4}$ \enskip\enskip
    \\
    \small
    \textsuperscript{1}Tianjin University \enskip
    \textsuperscript{2}The Hong Kong Polytechnic University \enskip 
    \textsuperscript{3}Shanghai Jiao Tong University \enskip 
    \textsuperscript{4}National University of Singapore \enskip \\
}

\newif\ifreview 
\newif\ifarxiv \newcommand{\arxiv}{\arxivtrue}
\newif\ifcamera 
\newif\ifrebuttal

\arxiv 

\pdfoutput=1
\documentclass[10pt,twocolumn,letterpaper]{article}

\usepackage{threeparttable}
\usepackage{times}
\usepackage{epsfig}
\usepackage{graphicx}
\usepackage{amsmath}
\usepackage{mathrsfs}
\usepackage{amssymb}
\usepackage{tabulary}
\usepackage{color}
\usepackage{multirow}
\usepackage{bm}
\usepackage{makecell}
\usepackage{bbding}
\usepackage{threeparttable}
\usepackage{color}\usepackage{graphicx}
\usepackage{amsmath}
\usepackage{amssymb}
\usepackage{booktabs}
\usepackage{multirow}
\usepackage{threeparttable}
\usepackage{times}
\usepackage{mathtools}
\usepackage{epsfig}
\usepackage{graphicx}
\usepackage{amsmath}
\usepackage{mathrsfs}
\usepackage{amssymb}
\usepackage{tabulary}
\usepackage{bm}
\usepackage{makecell}
\usepackage{bbding}
\usepackage{enumitem}
\usepackage{tabularx}
\usepackage{pifont} 
\usepackage{subcaption}

\usepackage{booktabs}

\newcommand{\cmark}{\ding{51}} 
\newcommand{\xmark}{\ding{55}} 

\ifreview \usepackage[review]{cvpr} \fi
\ifarxiv \usepackage[pagenumbers]{cvpr} \fi
\ifrebuttal \usepackage[rebuttal]{cvpr} \fi
\ifcamera \usepackage{cvpr} \fi

\usepackage{graphicx}
\usepackage{amsmath}
\usepackage{amssymb}
\usepackage{booktabs}

\usepackage{times}
\usepackage{microtype}
\usepackage{epsfig}
\usepackage[table,xcdraw]{xcolor}
\usepackage{caption}
\usepackage{float}
\usepackage{placeins}
\usepackage{color, colortbl}
\usepackage{stfloats}
\usepackage{enumitem}
\usepackage{tabularx}
\usepackage{xstring}
\usepackage{multirow}
\usepackage{xspace}
\usepackage{subcaption}
\usepackage{xcolor}
\usepackage[hang,flushmargin]{footmisc}

\ifcamera \usepackage[accsupp]{axessibility} \fi

% ...

\ifarxiv  \fi

\newcommand{\R}[1]{{%
    \textbf{%
        \ifstrequal{#1}{1}{\textcolor{red}{R#1}}{%
        \ifstrequal{#1}{2}{\textcolor{blue}{R#1}}{%
        \ifstrequal{#1}{3}{\textcolor{magenta}{R#1}}{%
        \ifstrequal{#1}{4}{\textcolor{teal}{R#1}}{%
                           \textcolor{cyan}{R#1}%
        }}}}%
    }%
}}

\usepackage{xr-hyper}

\makeatletter
\newcommand*{\addFileDependency}[1]{
  \typeout{(#1)}
  \@addtofilelist{#1}
  \IfFileExists{#1}{}{\typeout{No file #1.}}
}

\makeatother

\usepackage[pagebackref,breaklinks,colorlinks]{hyperref}
\usepackage[capitalize]{cleveref}
\crefname{section}{Sec.}{Secs.}
\crefname{table}{Table}{Tables}
\crefname{figure}{Fig.}{Figs.}

\begin{document}
\title{NoiseController: Towards Consistent Multi-view Video Generation via\\Noise Decomposition and Collaboration}
\author{\authorBlock}
\maketitle

\renewcommand{\thefootnote}{\fnsymbol{footnote}}
\renewcommand{\thefootnote}{\arabic{footnote}}

\begin{abstract}

High-quality video generation is crucial for many fields, including the film industry and autonomous driving. However, generating videos with spatiotemporal consistencies remains challenging. Current methods typically utilize attention mechanisms or modify noise to achieve consistent videos, neglecting global spatiotemporal information that could help ensure spatial and temporal consistency during video generation. In this paper, we propose the \textbf{\textit{NoiseController}}, consisting of \textbf{Multi-Level Noise Decomposition}, \textbf{Multi-Frame Noise Collaboration}, and \textbf{Joint Denoising}, to enhance spatiotemporal consistencies in video generation. In multi-level noise decomposition, we first decompose initial noises into scene-level foreground/background noises, capturing distinct motion properties to model multi-view foreground/background variations. Furthermore, each scene-level noise is further decomposed into individual-level shared and residual components. The shared noise preserves consistency, while the residual component maintains diversity.
In multi-frame noise collaboration, we introduce an inter-view spatiotemporal collaboration matrix and an intra-view impact collaboration matrix 
, which captures mutual cross-view effects and historical cross-frame impacts to enhance video quality. The joint denoising contains two parallel denoising U-Nets to remove each 
scene-level noise, 
mutually enhancing video generation.
We evaluate our \textbf{\textit{NoiseController}} on public datasets focusing on video generation and downstream tasks, demonstrating its state-of-the-art performance.


\end{abstract} 
\section{Introduction}


Video generation has provided tremendous visual data with realistic styles to train the computer vision models~\cite{li2022_bevformer,li2023_bevdepth,hu2021_fiery,philion2020_liftSplatShoot,wang2023exploring}. Multi-view video generation shows its unique challenges compared to other video generation tasks due to the additional need for cross-view consistency, which necessitates a more comprehensive understanding of the surrounding environment.


High consistency, controllability, and diversity are crucial for high-quality video generation. 
Achieving video consistency involves establishing spatiotemporal correspondences across multiple views while modeling interactions between foreground objects and background environments. Controllability refers to the ability to generate expected content. Recently, cross-view attention~\cite{zhao2024drivedreamer, wen2024panacea, li2023drivingdiffusion, yao2024mygo, gao2023magicdrive} and cross-frame attention mechanisms~\cite{li2023trackdiffusion, lu2023wovogen, wen2024panacea, wang2024driving, ma2024unleashinggeneralizationendtoendautonomous, zhang2024controlvideo} have been introduced in video diffusion models~\cite{singer2023makeavideo, blattmann2023align, bar2024lumiere}, making positive progress in generating diverse and consistent visual content. However, they still struggle to ensure accurate control and spatiotemporal consistency in video generation, leaving these challenges unaddressed.


The initial noise sequences significantly affect the final quality of the generated videos. Varying these sequences leads to different outputs, even using the same model and conditions. This indicates that the attention mechanisms employed in existing models lack an adequate constraint of the initial noise, leading to inconsistent starting points for denoising. Recent works alleviate this problem through noise reinitialization strategies, such as \textbf{noise modification}~\cite{mao2023guided, shirakawa2024noisecollage, shi2024dragdiffusion, mao2024theLottery} and \textbf{single-level noise decomposition}~\cite{10205416, ge2023preserve, Zhang_2024_CVPR, wu2023freeinit, ma2024unleashinggeneralizationendtoendautonomous, qiu2024freetraj}. Specifically, noise modification involves directly processing existing noise, such as editing noise images and rescheduling noise sequences. This method makes video generation more controllable, but directly modifying existing noise reduces video diversity and introduces artificial artifacts, which ultimately reduces realism. 
Single-level noise decomposition mitigates this problem by separating noise into shared and residual components. The shared noise can be selected either from fixed reference frames or by referencing the previous frame. However, this approach neglects global temporal information and multi-view foreground-background collaborations, which limits improvements in video quality (e.g., consistency, diversity, and controllability).
In this paper, we propose \textbf{\textit{NoiseController}}, a noise control framework equipped with \textbf{Multi-Level Noise Decomposition}, \textbf{Multi-Frame Noise Collaboration}, and \textbf{Joint Denoising} for consistent multi-view video generation. In multi-level noise decomposition, 
we first decompose the initial noises into the scene-level background and foreground noises, which capture distinct motion properties to model multi-view foreground/background variations. Then we employ individual-level decomposition to further decompose the scene-level noise into shared and residual components, enhancing multi-view video consistency and diversity. This two-level decomposition achieves accurate multi-view noise control, facilitating the modeling of noise collaborations.
In multi-frame noise collaboration, we utilize inter-view and intra-view collaborations to capture mutual cross-view effects and historical cross-frame impacts, enhancing multi-view video quality.
In the joint denoising stage, we leverage two parallel denoising U-Nets to separately remove the background and foreground noises. Notably, this design does not violate the principles of diffusion models. We follow recent diffusion-based works~\cite{gao2023magicdrive,li2023trackdiffusion} whose predicted noise is tolerable to non-standard normal distribution, to train denoising U-Nets.
Our design mimics the natural tendency of humans to allocate varying levels of attention to background and foreground elements when observing the world. Consequently, \textit{NoiseController} enables more controllable, consistent, and diverse video generation.

\textit{NoiseController} can generate multi-view spatial-temporal consistent videos, leading to state-of-the-art results in video generation tasks. 
Subsequently, these generated videos can significantly enhance the performance of downstream object detection and BEV segmentation.

\vspace{-0.05cm} 
\section{Related Work}

\subsection{Consistent Video Generation}

\begin{figure*}[t!]
\centering
\includegraphics[width=\linewidth]{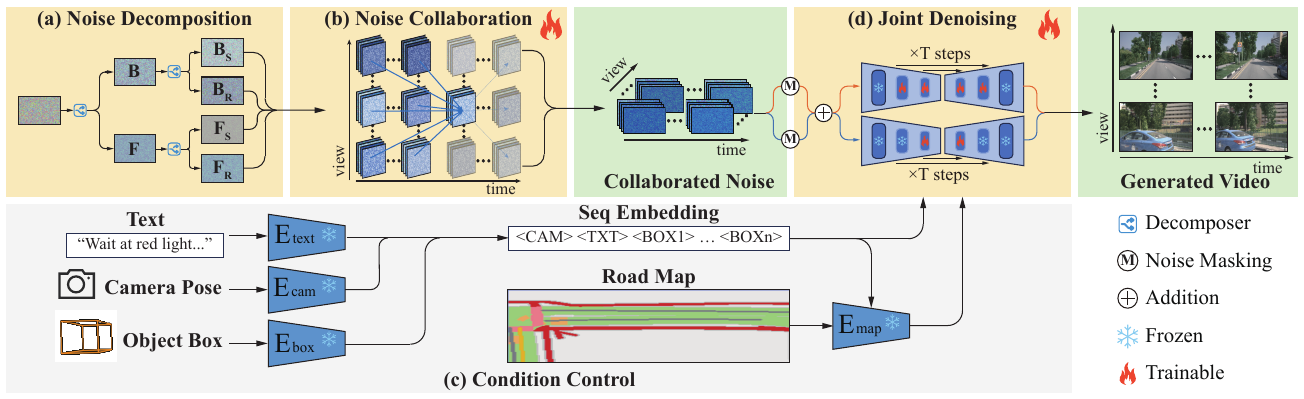}
\vspace{-0.28in}
\caption{Overview of \textbf{NoiseController}. In (a) multi-level noise decomposition, we use a decomposer to decompose the initial noises into scene-level background ($\bf B$) and foreground ($\bf F$) noises and individual-level noises by scaling with different coefficients, where each noise consists of individual-level shared ($\mathrm{S}$) and residual ($\mathrm{R}$) components. In (b) multi-frame noise collaboration, we respect the inter-view and intra-view collaboration matrices, yielding the shared components of scene-level noises for the following frames. The concatenation of the collaborated noise which is composed of shared components and randomly sampled residual noises, and the outputs of (c) condition control are fed into (d) joint denoising to predict $6$-view noises at $t^{th}$ denoising step.
}
\label{fig:overview}
\vspace{-0.2in}
\end{figure*}

Video generation has significantly addressed spatial and temporal consistency challenges, with various approaches emerging. We categorize these works into single- and multi-view video generation. Specifically, \textbf{single-view video generation} approaches~\cite{10205416, ge2023preserve, qiu2023freenoise, wu2023freeinit, zhou2024storydiffusion, wu2023tune, qiu2024freetraj, li2023trackdiffusion, Zhang_2024_CVPR} typically employ cross-frame attention mechanisms or modeling inter-frame motion to address temporal consistency.
Zhou et al.~\cite{zhou2024storydiffusion} ensure subject consistency through consistent self-attention within a batch of images. Wu et al.~\cite{wu2023tune} design intra-frame self-attention and cross-frame attention between the current frame and both the first and previous frames to ensure temporal consistency. Luo et al.~\cite{10205416} model motion in frame sequences by manipulating initial noise, using middle frame noise as the shared component across all frames, with independent residual as motion components to ensure background consistency.

\textbf{Multi-view video generation}~\cite{gao2023magicdrive, li2023drivingdiffusion, wang2024driving, wang2023drivedreamer, wen2024panacea, wen2024panaceapanoramiccontrollablevideo, lu2023wovogen, zhao2024drivedreamer, wu2024drivescape, ma2024unleashinggeneralizationendtoendautonomous, yao2024mygo} faces more challenging consistency,
as it requires maintaining both intra-view and cross-view consistency across frames. These works mainly ensure spatiotemporal consistency through two main methods: modeling cross-view and cross-frame relationships, or introducing additional conditions such as camera poses and motion trajectories. Wen et al.~\cite{wen2024panacea} propose decomposed 4D attention, using intra-view, cross-view, and cross-frame attention to maintain spatial and temporal consistency. Gao et al.~\cite{gao2023magicdrive} introduce various 3D geometric control conditions including road BEV maps, object bounding boxes, and camera poses to achieve fine-grained control, thereby enhancing spatiotemporal consistency. Huang et al.~\cite{huang2024subjectdrive} establish an external subject bank through external vehicle datasets to maintain foreground element consistency.

These works only consider local information when modeling video spatiotemporal relationships, such as adjacent views or frames. 
As information propagates along temporal or spatial dimensions, critical details may be forgotten, resulting in lacking global consistency. To address this challenge, our method models the cross-frame and cross-view relationships globally across all views and frames, ensuring global spatiotemporal consistency in the generated videos.

\subsection{Noise Modification}
Recently, many attempts~\cite{mao2023guided, shirakawa2024noisecollage, shi2024dragdiffusion, mao2024theLottery} have been conducted to explore the connection between image generation and initial noise, demonstrating that modifying the initial noise can improve image generation performance. Specifically, Mao et al.~\cite{mao2023guided} propose the fine-grained control without altering the denoising process by adjusting the generation tendency of initial noise images, laying the foundation for subsequent research. However, video generation faces more challenges due to the requisite consistency in spatial and temporal dimensions. Therefore, some works~\cite{wen2024panaceapanoramiccontrollablevideo, qiu2023freenoise, li2024tuning, wu2024motionbooth} improve video consistency through noise modification. Qiu et al.~\cite{qiu2023freenoise} reschedule a fixed set of noise frames through local shuffling, maintaining long-range correlations in noise sequences to ensure consistency in long videos while also introducing randomness to bring about content variations. Wen et al.~\cite{wen2024panaceapanoramiccontrollablevideo} propose a multi-view appearance noise prior, integrating the latent features of the first frame from multiple views into subsequent frames, further ensuring spatiotemporal consistency. 

These methods mainly enhance the temporal consistency of generated videos by controlling the overall noise. However, relying solely on such coarse-grained control through noise modification leads to homogenized video content and limits fine-grained controllability. In this paper, we propose a multi-level noise decomposition strategy, which enhances the frame-specific variations while ensuring consistency, thereby improving controllability.


\subsection{Noise Decomposition}
To alleviate the issues of content homogenization and limited fine-grained control mentioned above, recent works~\cite{10205416, ge2023preserve, Zhang_2024_CVPR, wu2023freeinit, ma2024unleashinggeneralizationendtoendautonomous, qiu2024freetraj} introduce a noise decomposition mechanism. This method breaks down complex noise into multiple sub-components, mitigating the denoising difficulties and improving video consistency. In detail, Luo et al.~\cite{10205416} propose a noise decomposition mechanism consisting of shared and residual noise with two jointly learned denoising networks to tackle content redundancy and temporal correlation, enhancing video coherence. Ge et al.~\cite{ge2023preserve} directly extend the i.i.d image noise to video noise prior, including mixed and progressive noise models containing shared and residual components to improve video generation performance via modeling inter-frame correlations. Based on image noise priors, Zhang et al.~\cite{Zhang_2024_CVPR} treat noise prediction as temporal residual learning, employing dual-path noise prediction to improve video consistency. Wu et al.~\cite{wu2023freeinit} explore differences in the frequency distribution of initial noise between training and inference, enhancing the temporal consistency primarily through low-frequency components in the denoising process.

These methods commonly leverage single-level noise decomposition to improve the quality of generated images or videos, neglecting the collaborations of cross-view noises. We introduce the multi-level noise decomposition, along with the inter-view spatiotemporal collaboration matrix and intra-view impact collaboration matrix to model scene-level and individual-level noise collaborations. It allows to fully integrate information from all perspectives across the entire time sequence, ensuring that the initial noise possesses spatiotemporal consistency.

\section{Method Overview}

\begin{figure}[t!]
\centering
\includegraphics[width=\linewidth]{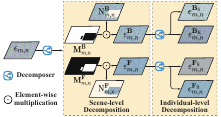}
\vspace{-0.3in}
\caption{Detailed architecture of multi-level noise decomposition. We decompose the initial noise $\epsilon_{m,n}$ into scene-level masked background noise $\bf N_{m,n}^{\bf B}$ and foreground noise $\bf N_{m,n}^{{\bf F}}$, following the distributions of $\mathcal{N}(\textbf{0}, {\bf I})$. We further decompose scene-level masked noises $\bf N_{m,n}^\mathbb{D}$ into individual-level masked shared components $\bf N_{m,n}^{\mathbb{D}_\mathrm{S}}$ and residual components $\bf N_{m,n}^{\mathbb{D}_\mathrm{R}}$.}
\label{fig:Decomposition}
\vspace{-0.2in}
\end{figure}

\vspace{0.05in}
\noindent{\bf Multi-Level Noise Decomposition}
As illustrated in Figure~\ref{fig:overview}(a), for the $m^{th}$ view at the $n^{th}$ frame, where $n \in {\{1, 2, \cdots, N\}}$ indicates the length of video frames, and $m \in \{1,2,..., 6\}$\footnote{6-view is a common setting in autonomous driving datasets.} indicates view index within the same frame, we propose a scene-level noise decomposition strategy, decomposing the noise $\epsilon_{m,n}$ into background noise $\epsilon_{m,n}^{\bf B}$ and foreground noise $\epsilon_{m,n}^{\bf F}$. We further introduce an individual-level noise decomposition strategy, decomposing scene-level background and foreground noises into shared components $\epsilon_{m, n}^{\mathbb{D}_{\mathrm{S}}}$ and residual components $\epsilon_{m, n}^{\mathbb{D}_{\mathrm{R}}}$, where $\mathbb{D} = \{\bf B, F \}$ 
(see Figure~\ref{fig:Decomposition}).


\vspace{0.05in}
\noindent{\bf Multi-Frame Noise Collaboration}
As illustrated in Figure~\ref{fig:overview}(b), we leverage the inter-view spatiotemporal collaboration matrix $\mathbb{S}= \{{\bf S}_n \in \mathbb{R}^{2 \times 6 \times 6 }|n=1,..., N \}$ with a sliding window whose length is $K$ to model 
the global cross-view cross-frame foreground-background variations. We resort to the intra-view impact collaboration matrix $\mathbb{I}=\{{\bf I}_k \in \mathbb{R}^{2 \times 2 \times 6 } |k=1, ..., K \}$ to learn intra-view collaborations between decomposed scene-level noises. We utilize the collaboration matrices $\mathbb{S}$ and $\mathbb{I}$ to compute the shared components $\epsilon_{m, n+1}^{\mathbb{D}_{\mathrm{S}}}$ of $\epsilon_{m, n+1}^{\mathbb{D}}$, which are then combined with the residual components $\epsilon_{m, n+1}^{\mathbb{D}_{\mathrm{R}}}$ to form the collaborated noise (see Figure~\ref{fig:Collaboration}).



\begin{figure*}[t!]
\centering
\includegraphics[width=\linewidth]{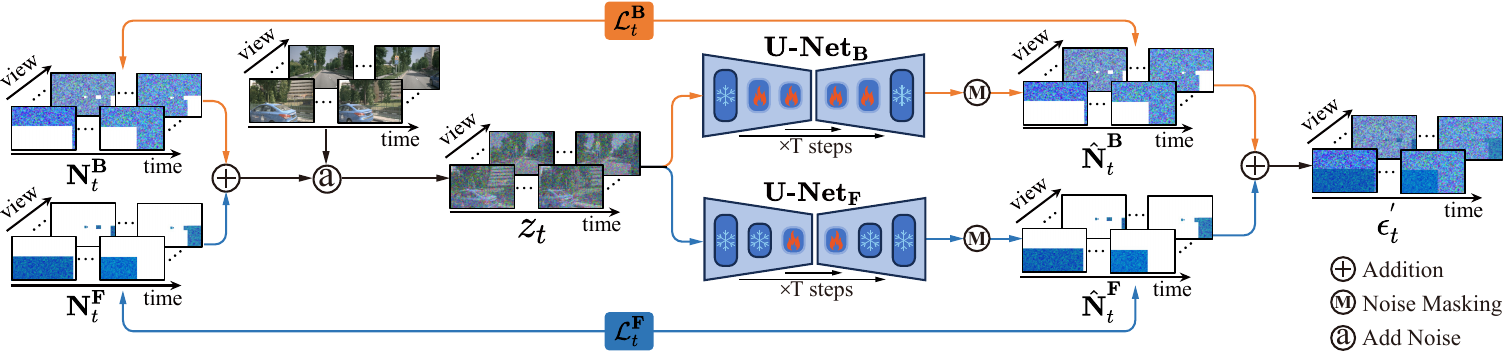}
\vspace{-0.25in}
\caption{Detailed architecture of joint denoising network. The $6$-view scene-level background and foreground noises $\epsilon_{t}^{\mathbb{D}}$ are masked by ${\bf M}^{\mathbb{D}}$ that is mapped from the 3D object box. The masked noises ${\bf N}_{t}^{\mathbb{D}}$ are added to the latent feature maps $x_0$ based on the spatial layout following SD~\cite{rombach2022high}. Taking inputs as the noisy latent feature maps $z_{t}$, two parallel denoising U-Nets are used to predict scene-level noises $\tilde{\epsilon}_{t}^{\mathbb D}$. Then we mask $\tilde{\epsilon}_{t}^{\mathbb D}$ by ${\bf M}_n^{\mathbb{D}}$ to obtain the predicted masked noises $\hat{\bf N}_{t}^{\mathbb{D}}$, which are finally combined to yield the final predicted noises $\epsilon_{t}^{'}$.}
\label{fig:JointDenoising}
\vspace{-0.2in}
\end{figure*}

\vspace{0.05in}
\noindent{\bf Joint Denoising}
As illustrated in Figure~\ref{fig:overview}(d), we first obtain the ground truth noises ${\bf N}_{t}^{\mathbb{D}}$ at the $t^{th}$ denoising step by element-wise multiplying the scene-level binary masks ${\bf M}^{\mathbb{D}} \in \mathbb{R}^{H \times W \times 3 \times 6\times N}$ ( with each element being 0 or 1) and the collaborated noises $\epsilon_{t}^{\mathbb{D}}$ (see Figure~\ref{fig:JointDenoising}). The latent feature maps $x_0$ and the ground truth noises ${\bf N}_{t}^{\mathbb{D}}$ are then combined to form the noised input. Two parallel denoising U-Nets, ${\bf \text{U-Net}}_{\mathrm{B}}$ and ${\bf \text{U-Net}}_{\mathrm{F}}$ process this noised input and predict noises ${\tilde{\epsilon}_{t}^{\mathbb D}}$. These predictions are subsequently masked with ${\bf M}^{\mathbb{D}}$ to produce the predicted masked noises $\hat{\bf N}_{t}^{\mathbb{D}}$, which are finally combined to obtain the final predicted noises $\epsilon_{t}^{'}$ at the $t^{th}$ denoising step.
 
\section{Architecture of NoiseController}
We introduce the detailed architectures of multi-level noise decomposition and multi-frame noise collaboration, and how to train our NoiseController for multi-view video generation in the following.
\subsection{Noise Controller}
\label{sec:NoiseControl}
\vspace{0.05in}
\noindent{\bf Multi-Level Noise Decomposition}
As illustrated in Figure~\ref{fig:Decomposition}, we propose a multi-level noise decomposition strategy to decompose the initial noise $\epsilon_{m,n}$ into different noise levels. 
To capture distinct motion properties for modeling multi-view foreground/background variations, we first propose a scene-level noise decomposition strategy, decomposing the initial noise $\epsilon_{m,n}$ into scene-level background noise $\epsilon_{m,n}^{\bf B}$ and foreground noise $\epsilon_{m,n}^{\bf F}$ for the $m^{th}$ view noise at the $n^{th}$ frame, formulated as:
\begin{equation}
    \begin{aligned}
        &\bf N_{m,n}^{\bf B} = \epsilon_{m,n}^{\bf B} \odot {\bf M}_{m,n}^{\bf B},~\bf N_{m,n}^{\bf F}=\epsilon_{m,n}^{\bf F} \odot {\bf M}_{m,n}^{\bf F}, \! \\
       & \epsilon_{m,n} \!=\! \bf N_{m,n}^{\bf B} \!\!+\!\!\bf N_{m,n}^{\bf F},\epsilon_{m,n}^{\bf B} \!\sim\! \mathcal{N}(\textbf{0}, \textbf{I}),\!
        \epsilon_{m,n}^{\bf F} \!\sim\! \mathcal{N}(\textbf{0}, \textbf{I}), 
    \end{aligned} 
\label{eq:share_residual}
\end{equation}
To enhance consistency and diversity in multi-view videos, we propose the individual-level decomposition, which further decomposes the scene-level noise $\epsilon_{m,n}^{\mathbb{D}}$ into shared components $\epsilon_{m, n}^{\mathbb{D}_{\mathrm{S}}}$ and residual components $\epsilon_{m, n}^{\mathbb{D}_{\mathrm{R}}}$ as:
\begin{equation}
\epsilon_{m, n}^{\mathbb{D}}=\epsilon_{m, n}^{\mathbb{D}_{\mathrm{S}}}+\epsilon_{m, n}^{\mathbb{D}_\mathrm{{R}}}
\end{equation}
The shared components of $6$-view scene-level noises in the first frame are randomly sampled, where $\epsilon_{m,1}^{{\bf B}_{\mathrm{S}}}$ and $\epsilon_{m,1}^{{\bf F}_{\mathrm{S}}}$ follow the Gaussian distribution of $\mathcal{N}({\bf 0}, \frac{\eta^{2}}{\eta^{2}+1} {\bf I})$ and $\mathcal{N}({\bf 0}, \frac{\lambda^{2}}{\lambda^{2}+1} {\bf I})$ respectively, 
helping to enhance the temporal consistency in the generated video frames.
The residual components $\epsilon_{m, n}^{\mathbb{D}_\mathrm{{R}}}$ are independently sampled for frame-specific variations, where each residual component follows a Gaussian distribution, i.e., $\epsilon_{m, n}^{{\bf B}_\mathrm{R}} \!\sim\!\ \mathcal{N}({\bf 0}, \frac{1}{\eta^{2}+1} {\bf I})$, $\epsilon_{m, n}^{{\bf F}_\mathrm{R}} \!\sim\!\ \mathcal{N}({\bf 0}, \frac{1}{\lambda^{2}+1} {\bf I})$, enhancing the diversity of the generated videos. This multi-level noise decomposition achieves accurate multi-view noise control.
\begin{figure}[t!]
\centering
\includegraphics[width=\linewidth]{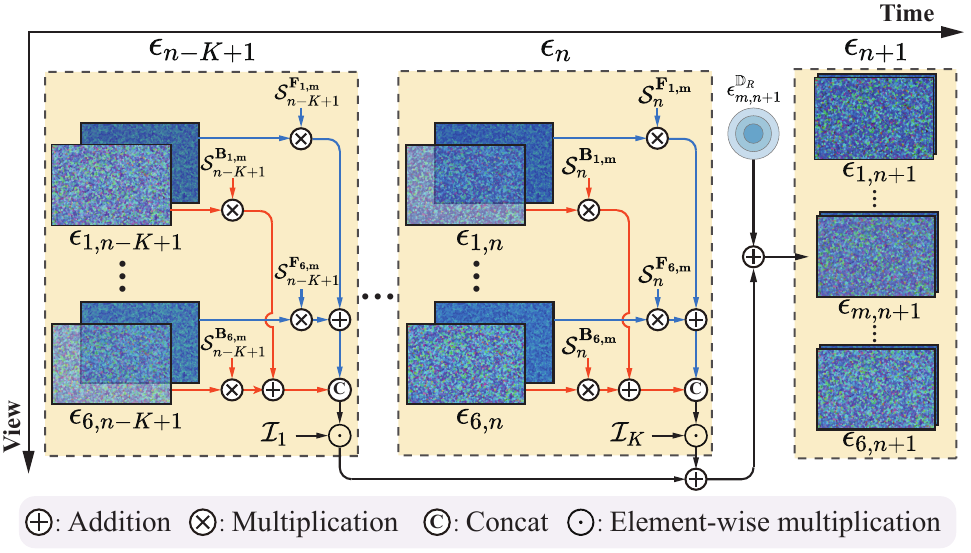}
\vspace{-0.3in}
\caption{Detailed architecture of multi-frame noise collaboration. We respect $6$-view noises of preceding $K$ frames to compute the shared components $\epsilon_{m, n+1}^{\mathbb{D}_{\mathrm{S}}}$. The concatenation of the product of $6$-view noises and inter-view spatiotemporal collaboration matrix is multiplied by the intra-view impact collaboration matrix. We sum the preceding $K$-frame collaborations to achieve shared components of scene-level noises $\epsilon_{m,n+1}^{\mathbb{D}_{\mathrm{S}}}$, which are then combined with sampled residual components $\epsilon_{m,n+1}^{\mathbb{D}_{\mathrm{R}}}$, yielding $6$-view noises at $(n+1)^{th}$ frame.}
\label{fig:Collaboration}
\vspace{-0.2in}
\end{figure}
\begin{figure}[t!]
\centering
\includegraphics[width=\linewidth]{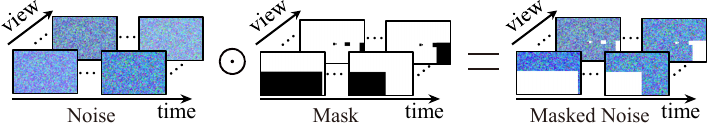}
\vspace{-0.28in}
\caption{The illustration of noise masking. We utilize scene-level $6$-view masks to mask background/foreground noises.
}
\label{fig:groundtruth}
\vspace{-0.2in}
\end{figure}
\vspace{0.05in}
\noindent{\bf Multi-Frame Noise Collaboration}
We propose the multi-frame noise collaboration to capture mutual cross-view effects and historical cross-frame impacts to enhance consistency in the initial noises. As illustrated in Figure~\ref{fig:Collaboration}, we respect the sliding window whose length is $K$ to enhance global cross-view temporal consistency during video generation. We leverage the inter-view spatiotemporal collaboration matrix $\mathbb{S}$ to model global inter-view collaborations of $6$-view noises based on preceding $K$ frames. The intra-view impact collaboration matrix $\mathbb{I}$ aims to model the local intra-view collaborations between each background and foreground noise within the same view. We define $\mathcal{I}_{m,k} \in \mathbb{R}^{2 \times 2}$ as the $k^{th}$ intra-view impact collaboration ${\bf I}_k$ within the sliding window at the $m^{th}$ view as follows:
\begin{equation}
\begin{aligned}
    & \mathcal{I}_{m,k}^{\bf B} = \begin{bmatrix}
        \mathcal{I}_{m,k}^{\bf B,B},~\mathcal{I}_{m,k}^{\bf F,B}
    \end{bmatrix}^{\top}, ~
     \mathcal{I}_{m,k}^{\bf F} = \begin{bmatrix}
        \mathcal{I}_{m,k}^{\bf B,F},~\mathcal{I}_{m,k}^{\bf F,F}
    \end{bmatrix}^{\top},  \\
    & \mathcal{I}_{m,k} =  \begin{bmatrix}
        \mathcal{I}_{m,k}^{\bf B}, \mathcal{I}_{m,k}^{\bf F}
    \end{bmatrix},~
    m \in {1,..., 6},
    \label{eq:impact}
\end{aligned}
\end{equation}
where $\mathcal{I}_{m,k}^{\bf B,B}$ represents the noise collaborations between scene-level noises within $m^{th}$ view. To calculate the collaborative contribution $\varepsilon_i^{\mathbb{D}}$ of $\epsilon^{\mathbb{D}}_{i}$ to the $6$-view noises $\epsilon_{n+1}$ at the $(n+1)^{th}$ frame, where $i=n-K+k$ represents the $i^{th}$ frame within the whole temporal sequence, we formulate the multi-frame noise collaboration as follows:
\begin{equation}
\begin{aligned}
    \varepsilon_{i}^{\mathbb{D}} = 
     \begin{bmatrix}
            \mathcal{S}_{i}^{\mathbb{D}_{1,1}} & \cdots & \mathcal{S}_{i}^{\mathbb{D}_{1,6}} \\
            \vdots & \ddots & \vdots \\
            \mathcal{S}_{i}^{\mathbb{D}_{6,1}} & \cdots & \mathcal{S}_{i}^{\mathbb{D}_{6,6}} \\
    \end{bmatrix}
    \begin{bmatrix}
        \epsilon_{1,i}^{\mathbb{D}}\\
        \vdots  \\
        \epsilon_{6,i}^{\mathbb{D}}\\
    \end{bmatrix}
    \odot
    \begin{bmatrix}
        \mathcal{I}_{1,k} \\
        \vdots \\
        \mathcal{I}_{6,k} \\
    \end{bmatrix},
    \label{eq:Matirx_Multiply}
\end{aligned}
\end{equation}
where $\mathcal{S}_{i}^{\mathbb{D}_{1,1}}$ represents the inter-view spatiotemporal collaboration between $\epsilon_{1,i}^\mathbb{D}$ and $\epsilon_{1,n+1}^\mathbb{D}$. We take the sum of $K$ collaborative contributions calculated using Eq.~\eqref{eq:Matirx_Multiply} as the $6$-view shared components $\epsilon_{n+1}^{\mathbb{D}_\mathrm{S}}$ for $6$-view scene-level noises $\epsilon_{n+1}^{\mathbb{D}}$ as follows:

\begin{equation}
\begin{aligned}
    \epsilon_{n+1}^{\mathbb{D}} = \sum_{i=n-K+1}^{n}{{\bf S}_{i} \cdot \epsilon_{i}^{\mathbb{D}} \odot {\bf I}_{k} + \epsilon_{n+1}^{\mathbb{D}_\mathrm{R}}},
\label{eq:noise_manipulation}
\end{aligned}
\end{equation}


\vspace{0.05in}
\subsection{Joint Denoising}

In Figure~\ref{fig:groundtruth}, for the $6$-view scene-level noises of the $N$ frame at $t^{th}$ denoising step, we leverage noise $\epsilon_{t}^{\mathbb{D}}$ and scene-level $6$-view $N$-frame masks ${\bf M}^{\mathbb{D}}$ to compute masked background noise ${\bf N}_{t}^{\bf{B}}$ and masked foreground noise ${\bf N}_{t}^{\bf{F}}$, obtaining ground-truth masked $6$-view $N$-frame noises $\epsilon_{t}$ as:
\begin{equation}
 \begin{aligned}
    {\bf N}_{t}^{\bf{B}} = \epsilon_{t}^{\bf B} \!\odot\! {\bf M}^{\bf B},~ 
    {\bf N}_{t}^{\bf{F}} = \epsilon_{t}^{\bf F} \!\odot\! {\bf M}^{\bf F},~
    \epsilon_{t} ={\bf N}_{t}^{\bf{B}} + {\bf N}_{t}^{\bf{F}},
     \end{aligned}
\end{equation}
where ${\bf M}^{\bf F} = 1 - {\bf M}^{\bf B}$ is the inverse value of ${\bf M}^{\bf B}$, representing the mask for foreground noise. $\odot$ represents the element-wise multiplication. 
Taking $z_{t}$ as input, which is obtained by adding the latent feature maps $x_0$ to the $t$-step masked noises, two parallel denoising U-Nets, ${\text{U-Net}_{\bf B}}$ and ${\text{U-Net}_{\bf F}}$, are employed to predict background noise $\tilde{\epsilon}_{t}^{\bf B}$ and foreground noise $\tilde{\epsilon}_{t}^{\bf F}$. We use ${\bf M}^{\mathbb{D}}$ to mask the background and foreground regions in predicted noises $\tilde{\epsilon}_{t}^{\mathbb D}$. Then, we can use the predicted masked background noises $\hat{\bf N}_{t}^{\bf{B}}$ and the predicted masked foreground noises $\hat{\bf N}_{t}^{\bf{F}}$ to compute the $\epsilon_{t}^{'}$ as follows:
\begin{equation}
    \begin{aligned}
     & \tilde{\epsilon}_{t}^{\bf B} = {\text {U-Net}}_{\bf B}(z_{t}),~\tilde{\epsilon}_{t}^{\bf F} = {\text {U-Net}}_{\bf F}(z_{t}), \\
      & \hat{\bf N}_{t}^{\bf{B}} \!=\! \tilde{\epsilon}_{t}^{\bf B} \!\odot\! {\bf M}^{\bf B},~ \hat{\bf N}_{t}^{\bf{F}} \!=\! \tilde{\epsilon}_{t}^{\bf F} \!\odot\! {\bf M}^{\bf F},~ \epsilon_{t}^{'} \!= \!\hat{\bf N}_{t}^{\bf{B}} + \hat{\bf N}_{t}^{\bf{F}},
    \end{aligned}
\label{eq:JointDenoise_2}
\end{equation}
where $\epsilon_{t}^{'}$ represents the predicted $6$-view $N$-frame holistic noises at denoising step $t$. 

\vspace{0.05in}
\subsection{Loss Function}
We initialize $\epsilon^{\bf B} \!\in\! \mathbb{R}^{2 \times 6 \times N}$ filled with $\frac{\eta^{2}}{\eta^{2}+1}$, and $\epsilon^{\bf F} \!\in\! \mathbb{R}^{2 \times 6 \times N}$ filled with $\frac{\lambda^{2}}{\lambda^{2}+1}$, yielding the coefficient of shared components. We leverage $L1$-norm ($|| \cdot ||_1$) to penalize the difference between the ground-truth and estimated coefficients of the shared components for the background as:
\begin{equation}
    \mathcal{L}^{\bf B} \!=\! ||\epsilon^{\bf{B}} \!-\!\!\! \sum_{i=n-K+1}^{n}{({\bf S}_{i} \cdot \epsilon_i^{\mathbb{D}} \!\odot\! {\bf I}^{\bf B}_k)\!\odot\! ({\bf S}_{i} \cdot \epsilon_i^{\mathbb{D}} \!\odot\! {\bf I}^{\bf B}_k)}||_1, \\
\label{eq:collaboration_loss}
\end{equation}
where ${\bf I}^{\bf B}_k \!=\! [{\mathcal{I}_{1,k}^{\bf B}},...,{\mathcal{I}_{6,k}^{\bf B}}]^\top$ is the collaborative impact for computing $\epsilon^{\bf{B}}_{n+1}$. Similarly, $\mathcal{L}^{\bf F}$ is computed by replacing background noises with foreground ones. We sum $\mathcal{L}^{\bf B}$ and $\mathcal{L}^{\bf B}$, yielding the holistic coefficient difference as: $\mathcal{L}^{C} \!=\! \mathcal{L}^{\bf B}\!+\!\mathcal{L}^{\bf F}$.
We minimize $L2$-norm ($|| \!\cdot\! ||_2$) to penalize the difference between ground-truth and estimated scene-level noises. The background noise difference is formulated as:
\begin{equation}
    \mathcal{L}_t^{\bf{B}_n} \!\!=\!\! \begin{cases}
        \!||{\bf M}_n^{\bf{B}} \!\odot\! ({\bf N}_{n,t}^{\bf{B}} - \hat{\bf N}_{n,t}^{\bf{B}})||_2 , n = 1, \\
        \!||{{\bf M}_n^{\bf{B}} \!\odot\! 
        ({\bf N}_{n,t}^{\bf{B}} \!\!-\!\! \sum_{i=max}^{n-1}}{{\bf S}_{i} \!\cdot\! \hat{\bf N}_{i,t}^{\bf{B}} \!\odot\! {\bf I}_{min}})||_2,
    \end{cases}   
\end{equation}
where $max = \max(n-K, 1)$, $min=i-max+1$. Similarly, $\mathcal{L}_t^{\bf{F}_n}$ is computed by replacing background noises with foreground ones. We define the final loss function as:
\begin{equation}
 \mathcal{L}^{\bf{B}}_t\!=\! \sum_{n=1}^{N}{\mathcal{L}_t^{\bf{B}_n}},~ 
 \mathcal{L}^{\bf{F}}_t\!=\! \sum_{n=1}^{N}{\mathcal{L}_t^{\bf{F}_n}},~
 \mathcal{L}^{} \!=\! \mathcal{L}^{C} + \mathcal{L}^{\bf{B}}_t + \mathcal{L}^{\bf{F}}_t.
\end{equation}
During training, we use the $\mathcal{L}^{}$ to optimize NoiseController at each denoising step $t$.

\section{Experiments}
\subsection{Implementation Details}
NoiseController utilizes pre-trained weights from MagicDrive (video)~\cite{gao2023magicdrive}, fine-tuning only newly added parameters and partial $\text{U-Net}$ layers to generate 16-frame videos. We employ data with a $224 \times 400$ resolution to train perception models like BEVFusion~\cite{liu2023bevfusion} and CVT~\cite{zhou2022cross}.
\vspace{-0.15in}
\begin{figure}[th!]
  \centering
  \begin{tabular}{@{\hspace{0mm}}c@{\hspace{1mm}}c@{\hspace{1mm}}c@{\hspace{1mm}}c}
  \includegraphics[width=0.46\linewidth]{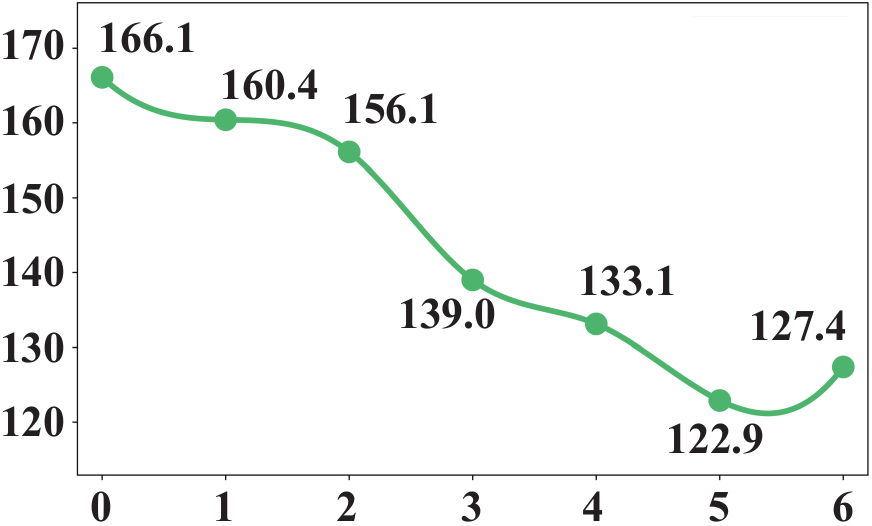}&
  \includegraphics[width=0.46\linewidth]{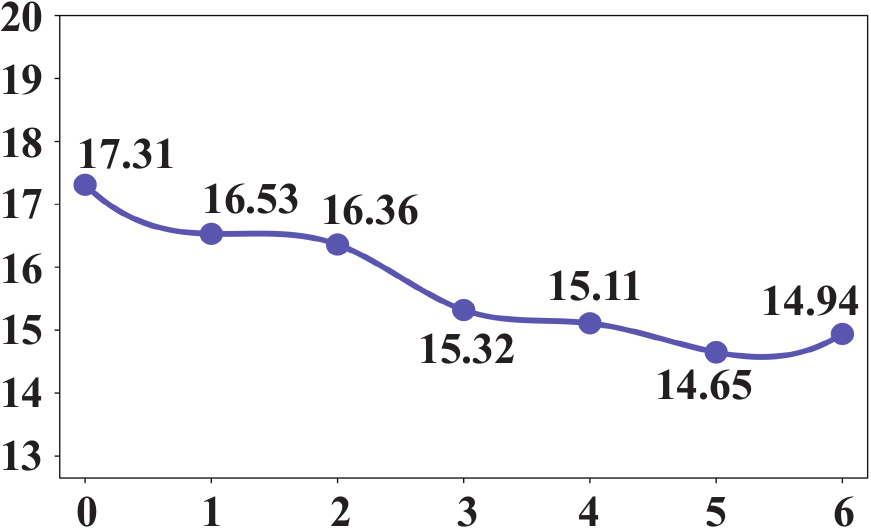}\\
  \vspace{-0.15in}
  \small{(a) FVD} & \small{(b) FID} 
  \end{tabular}
  \caption{Comparisons on the impact of sliding window length $K$. We report the performances in terms of FVD and FID on the validation set of nuScenes.}
  \label{fig:fvdfid}
  \vspace{-0.05in}
  \end{figure}
  
\begin{figure*}[t!]
\centering
 \includegraphics[width=0.9\linewidth]{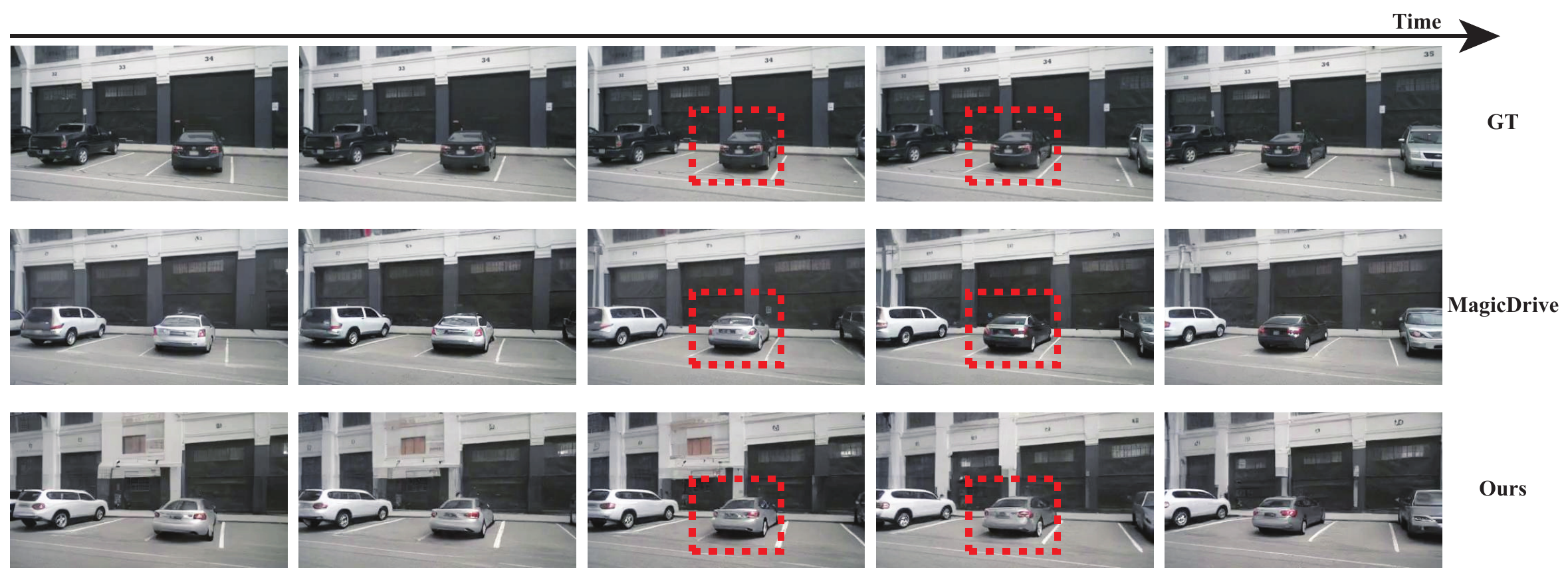}
\vspace{-0.1in}
\caption{The visualization comparison of cross-frame consistency between MagicDrive and NoiseController.}
\label{fig:cross_frame}
\vspace{-0.25in}
\end{figure*}

\subsection{Experimental Datasets}
We compare NoiseController with state-of-the-art methods on the nuScenes~\cite{caesar2020nuscenes} datasets, focusing on multi-view video generation. The dataset consists of 700 scenes for model training and 150 scenes for evaluation. We report the performances of multi-view video generation in terms of FID~\cite{heusel2017gans} and FVD~\cite{unterthiner2018towards}, where FID reflects the image quality and FVD measures the temporal consistency of videos.

We also evaluate NoiseController on perception tasks, including 3D object detection and BEV segmentation. Specifically, we utilize generated videos aligned with the annotations of nuScenes validation set to asses the performances of perception models, as a standard for evaluating controllability. Then, we generate videos to train the perception models to augment their performance. For the detection task, we use the mean Average Precision (mAP) and nuScenes Detection Score (NDS) as metrics, evaluated on BEVFusion~\cite{liu2023bevfusion}. We use mIoU as metrics for the segmentation task, evaluated on CVT~\cite{zhou2022cross}.

\subsection{Ablation Study of NoiseController}

\vspace{0.05in}
\noindent{\bf Sensitivity to Sliding Window Length}
We conduct experiments on the effect of sliding window length $K$ and report the performance
in terms of FVD and FID in Figure~\ref{fig:fvdfid}. We choose the window length $K$ from the set $\{0, 1,...,6\}$. With $K\!=\!0$, we disable the multi-frame noise collaboration. In this case, we randomly sample shared components $\epsilon_{n}^{\mathbb{D}_{\mathrm{S}}}$ and residual components $\epsilon_{n}^{\mathbb{D}_{\mathrm{R}}}$ of each scene-level noise $\epsilon_n^{\mathbb{D}}$ for the $n^{th}$ frame, achieving FVD $166.1$ and FID $17.31$.

With larger window length (e.g., $K\!\!=\!\!5$), we leverage the preceding $6$-view $K$-frame noises, yielding shared components $\epsilon_{n+1}^{\mathbb{D}_{\mathrm{S}}}$ for the $(n\!+\!1)^{th}$ frame. By capturing mutual cross-view effects and historical cross-frame impacts, we improve the multi-view video generation performances to FVD $122.9$ and FID $14.65$. Too large a window (e.g., $K\!=\!6$) requires extra computation to optimize the network during training, degrading performance under the same training steps. Hence, we set the default sliding window length to $K\!=\!5$ to balance computation and performance.

\setlength{\tabcolsep}{4.3pt}
\renewcommand{\arraystretch}{1.}
\begin{table}
\centering
\caption{Comparisons on the effectiveness of multi-level noise decomposition, multi-frame noise collaboration, and joint denoising network. We report the performances in terms of FVD and FID on the validation set of the nuScenes. \cmark and - represent the module is available and disabled respectively.}
\vspace{-0.3cm}
\footnotesize
\begin{threeparttable}
\begin{tabular}{ccc||cc}
\hline
\multicolumn{1}{c}{\textbf{Decomposition}} &  \multicolumn{1}{c}{\textbf{Collaboration}} &  \multicolumn{1}{c||}{\textbf{Joint Denoising}}
& \multicolumn{1}{c}{\textbf{FVD}$\downarrow $} & \multicolumn{1}{c}{\textbf{FID}$\downarrow $} 
 \\ \hline\hline
  -   &   -  &   - & 177.3 & 20.92   \\
  
-   &   -  &    \cmark   & 173.6 & 19.35  \\
 \cmark   & - & \cmark  &166.1 & 17.31  \\
 -   & \cmark & \cmark & 137.2 & 15.36   \\
 \cmark  & \cmark & \cmark & \bf{122.9} &\bf{14.65}  \\\hline
\end{tabular}
\end{threeparttable}
\label{tab:network_ablation}
\vspace{-0.15in}
\end{table}

\vspace{0.05in}
\noindent{\bf Analysis of Network Components} 
Multi-level noise decomposition, multi-frame noise collaboration, and joint denoising are three core modules of our NoiseController. We assess their effectiveness by disabling each of them, and report video generation performances in Table~\ref{tab:network_ablation}.
We first disable all three modules, NoiseController degrades into the baseline model, which neglects the global inter-view and local intra-view noise collaborations, achieving FVD $177.3$ and FID $20.92$ (see the first row of Table~\ref{tab:network_ablation}), which are significantly worse than NoiseController (last row). Then, we only reserve joint denoising, slightly improving the performances to FVD $173.6$ and FID $19.35$ (second row) by simple stacking models, demonstrating that integrating our proposed multi-level noise decomposition and multi-frame noise collaboration maximizes the effectiveness of joint denoising, achieving optimal performances.
Next, we disable the multi-frame noise collaboration of the NoiseController (third row). In this case, we decompose the initial noise into scene-level background and foreground noises consisting of further decomposed individual-level shared and residual components, where $6$-view shared components of the first frame and residual components of each frame are randomly sampled following the Gaussian distributions described in Section~\ref{sec:NoiseControl}. We choose the first frame as the base to calculate shared components as: $\epsilon_{m,n}^{\bf B_S} = \frac{\eta^2}{\eta^2+1}\epsilon_{m,1}, \epsilon_{m,n}^{\bf F_S} = \frac{\lambda^2}{\lambda^2+1}\epsilon_{m,1}$, for each following frame. It partially maintains temporal consistency. However, it fails to ensure consistency across different views, resulting in poor cross-view consistency.

We further solely disable multi-level noise decomposition of the NoiseController. In this case, we only employ single-level noise decomposition to decompose the initial noise into shared and residual noises. We leverage inter-view and intra-view matrices to model global and local noise collaborations, where both matrices solely focus on single-level decomposed shared and residual noises. This setting achieves FVD $137.2$ and FID $15.36$ (fourth row), which are worse than NoiseController due to its weak control over the background and foreground compared to the usage of multi-level noise decomposition.



\setlength{\tabcolsep}{1.5pt}
\renewcommand{\arraystretch}{1.}
\begin{table}[t]
    \centering
    \caption{We integrate NoiseController on TrackDiffusion and MagicDrive. Results are reported regarding FVD and FID on the validation set of YoutubeVIS~\cite{Yang2019vis} and nuScenes.}
    \vspace{-0.3cm}
    \footnotesize
    
    \begin{threeparttable}
    \begin{tabular}{c||c|c||c|c}
    \hline
     \multirow{1}{*}{\textbf{Method}} & \multirow{1}{*}{\textbf{Multi-View}} & \multirow{1}{*}{\textbf{Multi-Frame}}   &  \multirow{1}{*}{ \textbf{FVD}  $\downarrow$ }  &   \multirow{1}{*}{ \textbf{FID}  $\downarrow$ }   \\
      \hline\hline
    

    \textbf{TrackDiffusion}      &   & \cmark   &  605.0     &   41.81     \\
    \textbf{TrackDiffusion + ours} &    & \cmark    & \textbf{590.2} &     \bf{36.27}       \\ 
    \hline
    \textbf{MagicDrive}      & \cmark    & \cmark   & 177.3  &20.92      \\
    \textbf{MagicDrive + ours} & \cmark   &  \cmark    &\textbf{122.9}  & \bf{14.65}          \\\hline
    
    \end{tabular}
    \end{threeparttable}
    \label{tab:integrate}
    \vspace{-0.1in}
\end{table}

\vspace{0.05in}
\noindent{\bf Analysis of Method Integration Effectiveness} 
In Table~\ref{tab:integrate}, we report the performances of integrating our NoiseController on TrackDiffusion and MagicDrive. NoiseController outperforms the baseline model (MagicDrive) in terms of FVD and FID, achieving significant improvements of 30.7\% and 30.0\%, respectively. NoiseController also surpasses the baseline model (TrackDiffusion~\cite{li2023trackdiffusion}). 
This is because multi-level noise decomposition captures distinct motion properties, which helps to model multi-view foreground/background variations. Furthermore, it achieves accurate multi-view noise control, which provides a chance for multi-frame noise collaboration to capture mutual cross-view effects and historical cross-frame impacts, ultimately leading to performance improvements.
\setlength{\tabcolsep}{6.8pt}
\renewcommand{\arraystretch}{1.}
\begin{table}
\centering
\caption{We evaluate the effectiveness of inter-view spatiotemporal collaboration matrix $\mathbb{S}$ and intra-view impact collaboration matrix $\mathbb{I}$ on the nuScenes validation set. \cmark, \xmark, and - represent the matrix is learnable, disabled, and fixed, respectively.
}
\vspace{-0.3cm}
\footnotesize
\begin{threeparttable}
\begin{tabular}{cc||cc}
\hline
\multicolumn{1}{c}{{\bf Spatiotemporal Matrix} $\mathbb{S}$} &  \multicolumn{1}{c||}{{\bf Impact Matrix} $\mathbb{I}$}
& \multicolumn{1}{c}{\textbf{FVD}$\downarrow $} & \multicolumn{1}{c}{\textbf{FID}$\downarrow $} 
 \\ \hline\hline
 \xmark   & \cmark &154.4 & 17.26   \\
 -   & \cmark & 183.4& 22.71   \\
 \cmark   & \xmark &128.1 &15.15  \\
 \cmark   & - & 135.8& 15.29  \\
 \cmark  & \cmark &\bf{122.9}& \bf{14.65}   \\\hline
\end{tabular}
\end{threeparttable}
\label{tab:matrix}
\vspace{-0.10in}
\end{table}

\vspace{0.05in}
\noindent{\bf Analysis of Multi-Frame Noise Collaboration Effectiveness} 
We examine the effectiveness when two collaboration matrixes are disabled, fixed or learnable in terms of FVD and FID in Table~\ref{tab:matrix}. We fill the matrix with ones if it is fixed, representing noise collaboration of each view or scene-level noise is the same. 
First, we disable $\mathbb{S}$ and set $\mathbb{I}$ to be learnable, the performances degrade significantly (see the first row of Table~\ref{tab:matrix}), indicating that $\mathbb{S}$ is essential for modeling cross-view temporal consistency. Next, we disable $\mathbb{I}$ and set $\mathbb{S}$ to be learnable, and the performances also degrade (third row), demonstrating that $\mathbb{I}$ is important for capturing historical cross-frame impacts. Moreover, fixing either $\mathbb{S}$ or $\mathbb{I}$ results in the worst performances (second row and fourth row), because using the same $6$-view noise collaborations across frames fails to account for the fact that not all views or all frames are equally related (e.g., the front left view may differ significantly from the back right view, and the first frame may be less relevant to the last frame).

\begin{figure}[t!]
\centering
\includegraphics[width=1\linewidth]{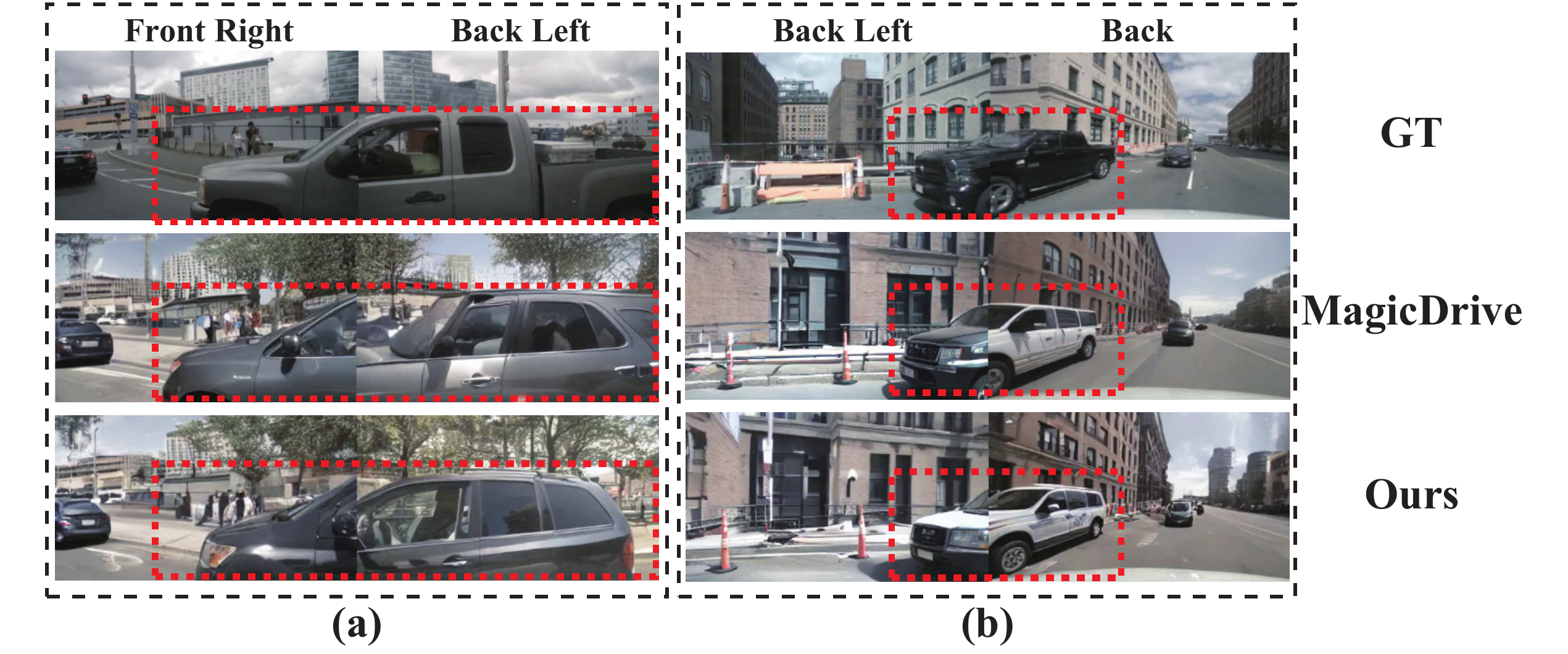}
\vspace{-0.3in}
\caption{The visualization comparison of cross-view consistency between MagicDrive and NoiseController.}
\label{fig:cross_view}
\vspace{-0.1in}
\end{figure}

\begin{figure}[t!]
\centering
\includegraphics[width=1\linewidth]{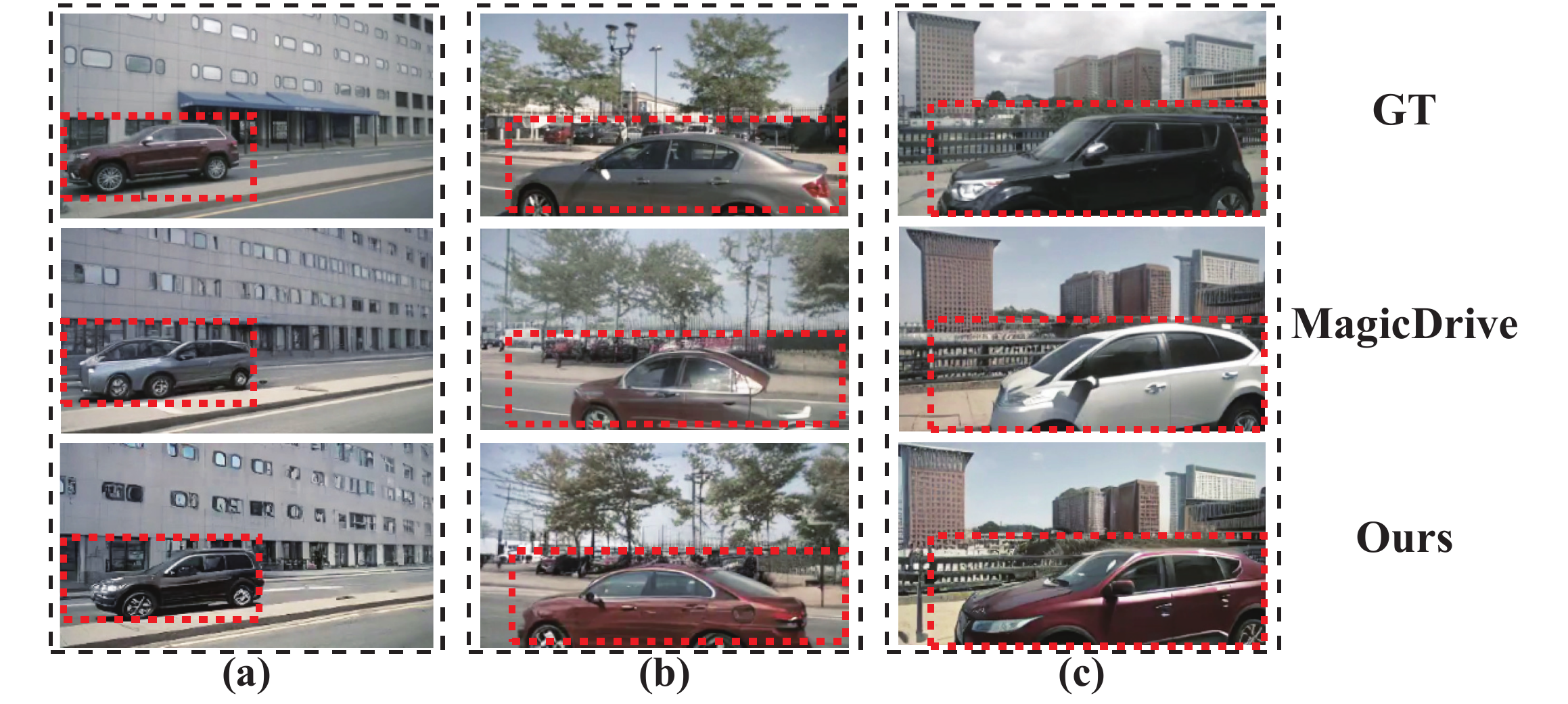}
\vspace{-0.3in}
\caption{The visualization comparison of foreground details. (a), (b) and (c) are three examples.}
\label{fig:fg_gen}
\vspace{-0.2in}
\end{figure}

\begin{figure*}[h!]
\centering
\includegraphics[width=0.95\linewidth]{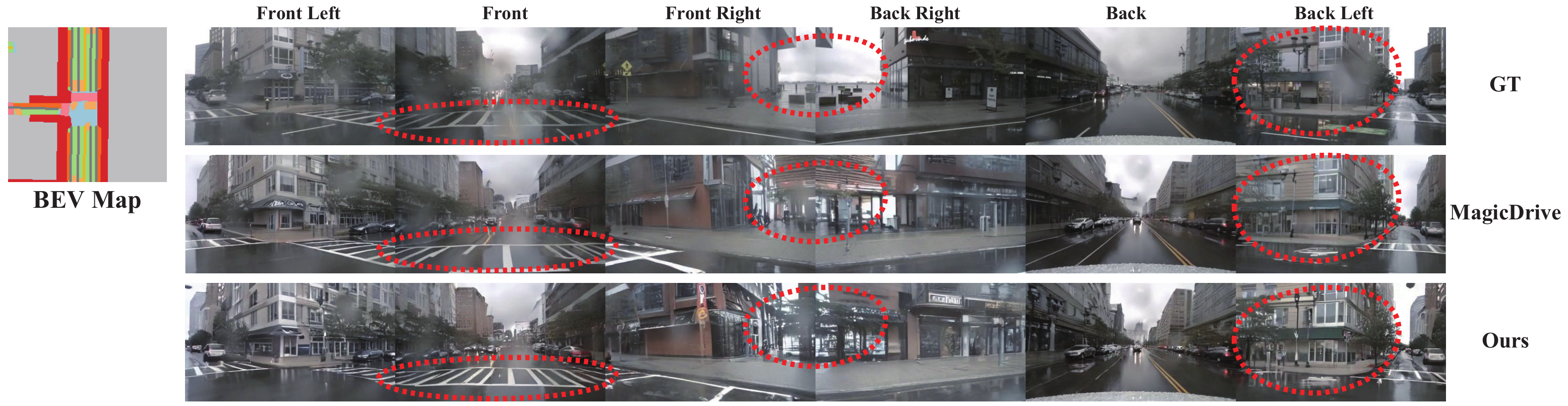}
\vspace{-0.15in}
\caption{The visualization comparison of background details.}
\label{fig:bg_gen}
\vspace{-0.2in}
\end{figure*}

\subsection{Main Results}

\setlength{\tabcolsep}{1.9pt}
\renewcommand{\arraystretch}{1.}
\begin{table}[t]
    \centering
    \caption{We compare NoiseController with state-of-the-art multi-view methods on the validation set of the nuScenes.}
    \vspace{-0.35cm}
    \footnotesize
    \begin{threeparttable}
    \begin{tabular}{c||c|c||c|c}
    \hline
     \multirow{1}{*}{\textbf{Method}}  & \multirow{1}{*}{\textbf{Multi-Frame}}& \multirow{1}{*}{\textbf{Resolution}}   &  \multirow{1}{*}{ \textbf{FVD}  $\downarrow$ }  &   \multirow{1}{*}{ \textbf{FID}  $\downarrow$ }   \\
      \hline\hline
    
    BEVGen~\cite{swerdlow2024street}       &  &224*400  &  -     &    25.54      \\
    
    BEVControl~\cite{yang2023bevcontrol}       &  & 512*512 &  -     &  24.85      \\
    MagicDrive (img)~\cite{gao2023magicdrive}    &    & 224*400   & - &  16.20    \\
    WoVoGen~\cite{lu2023wovogen}     & \cmark & 256*448   &417.7 & 27.60            \\ 
    DriveDreamer~\cite{wang2023drivedreamer}    &  \cmark  & 256*448 &340.8  & 14.90          \\
    MagicDrive (video)~\cite{gao2023magicdrive}    &  \cmark & 224*400  &177.3  & 20.92          \\
    Panacea~\cite{wen2024panacea}    &  \cmark   & 256*512&139.0  & 16.96          \\
    
    SubjectDrive~\cite{huang2024subjectdrive} & \cmark   &  256*512    &124  & 15.96          \\
     Panacea+~\cite{wen2024panaceapanoramiccontrollablevideo} & \cmark & 512*1024 & 103 & 15.5 \\
    MagicDriveDiT~\cite{gao2024magicdrivedit} & \cmark & 848*1600 & 94.84 & - \\
    \hline\hline
    \textbf{NoiseController}    &  \cmark & 224*400 & \bf{122.9} &  \bf{14.65}         \\
    \textbf{NoiseController*}    &  \cmark & 512*1024 & \bf{87.23} &  \bf{13.72}         \\
    \hline
    
    \end{tabular}
    \end{threeparttable}
    \vspace{-0.1in}
    \label{tab:sota}
\end{table}

\vspace{0.05in}
\noindent{\bf Image Quality and Temporal Consistency}
In Table~\ref{tab:sota}, we measure the image quality and temporal consistency of the image/video generated by various methods in terms of FID and FVD. NoiseController achieves 14.65 FID and 122.9 FVD, which is competitive to other lower-resolution methods. Additionally, we extend our method to higher-resolution generation, yielding 13.72 FID and 87.23 FVD, surpassing other methods. We show visualization of generated videos in Figures~\ref{fig:cross_frame} and \ref{fig:cross_view}, demonstrating that NoiseController maintains better spatiotemporal consistency. 

\setlength{\tabcolsep}{0.9pt}
\renewcommand{\arraystretch}{1.}
\begin{table}[t]
\centering
\caption{We compare generation fidelity with stree-view generation methods. For each task, we use our generated images aligned with the nuScenes validation set annotations to test the models trained on its training set. We report the performances of the detection model (BEVFusion) and the segmentation model (CVT).
}
\vspace{-0.32cm}
\footnotesize
\begin{threeparttable}
\begin{tabular}{c||cc|c||c}
\hline
 \multirow{3}{*}{\textbf{Method}}    &  \multicolumn{3}{c||}{\textbf{Foreground Control}}  & \multicolumn{1}{c}{\textbf{Background Control}}  \\
 \cline{2-5}
 
 & \multicolumn{2}{c|}{\textbf{Detection}}  & \multicolumn{1}{c||}{\textbf{Segmentation}}  & \multicolumn{1}{c}{\textbf{Segmentation}} \\
 \cline{2-5}

& \multicolumn{1}{c}{$ \text{mAP} \uparrow $} & \multicolumn{1}{c|}{$ \text{NDS} \uparrow $} 
& \multicolumn{1}{c||}{$ \text{Vehicle IoU} \uparrow $}
& \multicolumn{1}{c}{$ \text{Road IoU} \uparrow $}
 \\ \hline\hline
Reference-score   & 35.54 & 41.21 & 33.66  &72.21 \\
\hline
BEVGen   &- &-  & 5.89 &50.20 \\
BEVControl  & -  & -    & 26.80  & 60.80   \\
MagicDrive (img)    &12.30 &23.32  & 27.01 & 61.05  \\
\hline\hline
{\textbf{NoiseController}}   &\textbf{20.93} & \textbf{27.96}  & \textbf{27.32} &\textbf{64.85}   \\ 
 
  \hline
\end{tabular}
\end{threeparttable}

\label{tab:control}
\vspace{-0.15in}
\end{table}

\setlength{\tabcolsep}{3.3pt}
\renewcommand{\arraystretch}{1.}
\begin{table}[t]
    \caption{Application of using the generated data to augment the downstream tasks. We compare the data augmentation effect on the detection model (BEVFusion) and the segmentation model (CVT). The results are evaluated on nuScenes validation set.}
    \vspace{-0.35cm}
    \centering
        \centering
        \footnotesize
        \begin{threeparttable}
        \begin{tabular}{c||cc||cc}
        \hline
        \multirow{2}{*}{\textbf{Data}} & \multicolumn{2}{c||}{\textbf{Detection} } & \multicolumn{2}{c}{\textbf{Segmentation} } \\
        \cline{2-5}
        &\multirow{1}{*}{$ \text{mAP} \uparrow $} &\multirow{1}{*}{$ \text{NDS} \uparrow $}  &\multirow{1}{*}{$ \text{Vehicle IoU} \uparrow $} & \multirow{1}{*}{$ \text{Road IoU} \uparrow $}  \\ \hline \hline 
        w/o aug. &32.48    &37.61 &36.00    &74.30   \\ \hline 
       w/ BEVGen &34.93 &38.77  &36.60    &71.90   \\
       {w/ MagicDrive (img)} &35.14    &39.63      &40.34    &79.56    \\ 
       {w/ DriveDreamer} & 35.80 & 39.50 &  -  &  - \\
        {w/ DrivingDiffusion} & - & {43.40}  &  - & - \\\hline\hline 
        \textbf{w/ NoiseController}   &\textbf{36.07}    &{\textbf{43.54} }      &{\textbf{40.97}}    &{\textbf{80.13}}   
        \\\hline 
        \end{tabular}
        \end{threeparttable}
    \label{tab:downstream}
    \vspace{-0.17in}
\end{table}

\vspace{0.05in}
\noindent{\bf Foreground and Background Controllability}
As shown in Table~\ref{tab:control}, we measure the controllability of various methods on foreground and background objects. We use BEVFusion to detect the foreground objects, and CVT to segment both foreground and background. NoiseController provides high-quality data that significantly improves the performance of downstream detection and segmentation models. This is because our multi-level noise decomposition strategy enhances the consistency of both foreground and background through accurate control. Furthermore, our multi-frame noise collaboration effectively models the spatial and temporal relations to capture mutual cross-view effects and historical cross-frame impacts, resulting in better consistency and more details. Figure~\ref{fig:fg_gen} and Figure~\ref{fig:bg_gen} demonstrate that NoiseController generates high-quality foreground objects and background scenes.

\vspace{0.05in}
\noindent{\bf Comparison on Downstream Applications}
In Table~\ref{tab:downstream}, we evaluate the effectiveness of our method in producing data to augment the perception tasks. Specifically, we train the BEVFusion model in Camera-Only settings with NoiseController's generated data for object detection. We utilize the generated data from NoiseController to train CVT for BEV segmentation. By integrating the data generated by NoiseController, we significantly improve detection and segmentation tasks. We fully fuse global and local information, generating diverse and consistent videos. More details can be found in Sec.5 of the supplementary file.
\section{Conclusion}


In this paper, we propose a noise control model for multi-view video generation, NoiseController, which focuses on enhancing spatial and temporal consistencies. We model the cross-view and cross-frame relations through multi-leve noise decomposition and collaboration. It allows a full fusion of global and local information, achieving consistent initial noises. Additionally, we utilize joint denoising, focusing on the background and foreground separately, to generate consistent videos from the consistent initial noises. NoiseController achieves state-of-the-art performance in video generation and downstream tasks.

{\small
\bibliographystyle{ieee_fullname}
\bibliography{11_references}
}


\end{document}


\title{\paperTitle \\ Supplemental Material}
\author{\authorBlock}
\maketitle

\appendix
\section*{Appendix}
\label{sec:appendix}
In this supplemental material, we first provide more implementation and training details, and then present more results.

\section{Implementation and training details}
The number of view(e.g. $V$) is 6 in the NuScenes dataset~\cite{caesar2020_nuscenes} , arranged in clock-wise direction starting from cam\_front view. In the foreground condition, the orientation of bounding boxes is defined in the camera coordinate system as shown in Figure~\ref{fig:orientation_coords_sys}. 

\begin{figure}[t!]
\centering
\includegraphics[width=\linewidth]{figs/method/box_orientation_sys_2.pdf}
\vspace{-0.1in}
\caption{Box orientation coordinate system. }
\label{fig:orientation_coords_sys}
\vspace{-0.15in}
\end{figure}

\begin{figure*}[t!]
\centering
\includegraphics[width=\linewidth]{figs/quanlitative_results/fg_bg_select/fg-bg-0046092508b14f40a86760d11f9896bb.pdf}
\vspace{-0.1in}
\caption{The visualization of foreground controlling generation. }
\label{fig:foreground_control_visual}
\vspace{0.2in}
\end{figure*}

\begin{figure*}[t!]
\centering
\includegraphics[width=\linewidth]{figs/quanlitative_results/fg_bg_select/fg-80e281bf369a4f849efaa5a9052cbd9b.pdf}
\vspace{-0.1in}
\caption{The visualization of foreground controlling generation. }
\label{fig:foreground_control_visual}
\vspace{0.2in}
\end{figure*}

\begin{figure*}[t!]
\centering
\includegraphics[width=\linewidth]{figs/quanlitative_results/fg_bg_select/bg-7ac3bb7fba5c4852a685555407cd10f1.pdf}
\vspace{-0.1in}
\caption{The visualization of background controlling generation.}
\label{fig:foreground_control_visual}
\vspace{0.2in}
\end{figure*}

\begin{figure*}[t!]
\centering
\includegraphics[width=\linewidth]{figs/quanlitative_results/fg_bg_select/bg-a4a9d61254d148fba35d53277b5246f8.pdf}
\vspace{-0.1in}
\caption{The visualization of background controlling generation.}
\label{fig:foreground_control_visual}
\vspace{0.2in}
\end{figure*}

\begin{figure*}[t!]
\centering
\includegraphics[width=\linewidth]{figs/quanlitative_results/sketch/sketch-fg_180_170_50_-20.pdf}
\vspace{-0.1in}
\caption{The visualization of foreground controlling generation. }
\label{fig:foreground_control_visual}
\vspace{0.2in}
\end{figure*}

\begin{figure*}[t!]
\centering
\includegraphics[width=\linewidth]{figs/quanlitative_results/sketch/sketch-bg125.pdf}
\vspace{-0.1in}
\caption{The visualization of background controlling generation. }
\label{fig:foreground_control_visual}
\vspace{0.2in}
\end{figure*}

{\small
\bibliographystyle{ieee_fullname}
\bibliography{11_references}
}